\documentclass[conference]{IEEEtran}
\IEEEoverridecommandlockouts
\usepackage[english]{babel}
\usepackage[autostyle, english = american]{csquotes}
\usepackage{cite}
\usepackage{amsmath,amsthm,amssymb,amsfonts}
\usepackage{algorithmic}
\usepackage{graphicx}
\usepackage{textcomp}
\usepackage{xcolor}
\def\BibTeX{{\rm B\kern-.05em{\sc i\kern-.025em b}\kern-.08em
    T\kern-.1667em\lower.7ex\hbox{E}\kern-.125emX}}

\usepackage{amsmath,amsfonts,bm}

\usepackage{amsthm}       % For theorem environments
\usepackage{thmtools}     % For restatable support

\newcommand{\bp}{\begin{pmatrix}}
\newcommand{\ep}{\end{pmatrix}}

\def\eqref#1{equation~\ref{#1}}
\def\1{\bm{1}}

\def\vf{{\bm{f}}}

\def\vw{{\bm{w}}}
\def\vx{{\bm{x}}}

\def\mH{{\bm{H}}}

\def\mM{{\bm{M}}}

\def\mW{{\bm{W}}}
\def\mX{{\bm{X}}}

\DeclareMathAlphabet{\mathsfit}{\encodingdefault}{\sfdefault}{m}{sl}
\SetMathAlphabet{\mathsfit}{bold}{\encodingdefault}{\sfdefault}{bx}{n}

\def\gC{{\mathcal{C}}}

\def\gE{{\mathcal{E}}}
\def\gF{{\mathcal{F}}}
\def\gG{{\mathcal{G}}}

\def\gI{{\mathcal{I}}}

\def\gV{{\mathcal{V}}}

\def\sR{{\mathbb{R}}}

\newcommand{\tbf}[1]{\textbf{#1}}

\usepackage{comment}
\usepackage{hyperref}
\usepackage{url}
\usepackage{booktabs,siunitx}
\begin{document}

\title{
Boltzmann Graph Ensemble Embeddings for Aptamer Libraries
% Graph Ensemble Embeddings for Aptamer Discovery
% Physics-Informed Graph Embeddings for Anomaly Screening and Robust Library Design of Aptamers
% Physics-Informed Graph Embeddings for Robust Community Detection and Aptamer Library Design
% Graph Techniques for Aptamer Community Detection and Anomaly Screening
% Subgraph Fingerprint Statistics for
% Interactive Binding
% via Graph Techniques
% Graph Techniques for Detecting Anomalous and Adversarial Competitors in Iterative Binding Selection
% Graph Techniques for Competitive Anomaly Detection in SELEX
% Graph Techniques for Detecting Anomalous Competitors in SELEX
% Statistical Ensembles of Subgraph Fingerprints for
% Anomalous Aptamer Detection
% {\em de novo} Aptamer Design\\
% {
% \footnotesize
% \textsuperscript{*}Note: Sub-titles are not captured for https://ieeexplore.ieee.org  and
% should not be used}
\thanks{
%{\em Removed for double blind review.}\hfill
%\vspace{10ex} \phantom{
\noindent
$^*$ equal contribution sorted alphabetically.\\
This research is partially supported by the STEM scholars program from the Glenn W. Bailey Foundation, under grant number number 0633-001.  This research is also supported by Simons Foundation Grant number 510776.
\\
Correspond to justin@math.ucla.edu
\\
{$\dagger$ - member, IEEE
}\\
{$\ddagger$ - Departments 
Chemistry \& Biochemistry, Bioengineering,
Semel Institute for Neuroscience and Human Behavior, the
Hatos Center for Neuropharmacology,
 and the California NanoSystems Institute
}
} % phantom
% \end{comment}
}

\author{
 \IEEEauthorblockN{Starlika Bauskar$^*$}
 \IEEEauthorblockA{
 \textit{Texas Tech University}\\
 Lubbock, USA% \\
 }
 \and
 \IEEEauthorblockN{Jade Jiao$^*$}
 \IEEEauthorblockA{
 \textit{Pomona College}\\
 Claremont, USA %\\
 }
 \and
 \IEEEauthorblockN{Narayanan Kannan$^*$}
 \IEEEauthorblockA{%\textit{dept. name of organization (of Aff.)} \\
 \textit{University of California, Los Angeles}\\
 Los Angeles, USA \\
 }
 \and
 \IEEEauthorblockN{Alexander Kimm$^*$}
 \IEEEauthorblockA{
 \textit{University of California, Irvine}\\
 Irvine, USA %\\
 }
 \and
 \IEEEauthorblockN{
 Justin M. Baker, Matthew J. Tyler, Andrea L. Bertozzi$^\dagger$}
 \IEEEauthorblockA{\textit{Dept. of Mathematics and California NanoSystems Institute} \\
 \textit{University of California, Los Angeles}\\
 Los Angeles, USA
 }
 \and
 \IEEEauthorblockN{ Anne M. Andrews$^{\ddagger}$}
 \IEEEauthorblockA{\textit{
 {Depts. of Psychiatry \& Biobehavioral Sciences
 }}
 \\
 \textit{University of California, Los Angeles}\\
 Los Angeles, USA
 }
 }
%\IEEEauthorblockN{Anonymous Authors}
%\vspace{8ex}
%}

\maketitle

\begin{abstract}
Machine-learning methods in biochemistry commonly represent molecules as graphs of pairwise intermolecular interactions for property and structure predictions.
Most methods operate on a single graph, typically the minimal free energy (MFE) structure, for low-energy ensembles (conformations) representative of structures at thermodynamic equilibrium.
We introduce a thermodynamically parameterized exponential‑family random graph (ERGM) embedding that models molecules as Boltzmann‑weighted ensembles of interaction graphs.
We evaluate this embedding on SELEX datasets, where experimental biases (e.g., PCR amplification or sequencing noise) can obscure true aptamer–ligand affinity, producing anomalous candidates whose observed abundance diverges from their actual binding strength.
% We evaluate this embedding on SELEX datasets, where experimental bias can obscure true aptamer–ligand binding affinity resulting in anomalous data.
% We consider the dynamics‑driven task of determining aptamer–ligand binding affinity using experimental SELEX datasets where experimental bias results in some anomalous data.
% \tcb{Maybe like this: We evaluate this embedding on the prediction of aptamer–ligand binding outcomes using SELEX datasets that exhibit anomalies from experimental bias, relying on ensembles of low-energy unbound aptamer conformations. }\tcb{I think we just need to change the dynamics-driven part. But I thought the problem is we're talking directly about the binding state and not making it clear that we're doing structure analysis on the unbound state. I thought professor Andrews was saying we don't implement a (sub) structural dynamics-driven solution. Ohh ok. But wait doesn't that mean that we shouldn't use "aptamer ligand binding" directly if we are only evaluating low energy conformations of unbound aptamers?} 
We show that the proposed embedding enables robust community detection and subgraph‑level explanations for aptamer-ligand affinity, even in the presence of biased observations.
This approach may be used to identify low-abundance aptamer candidates for further experimental evaluation.
% Recent advances in machine learning for biochemistry utilize graph embeddings of pairwise interactions to accurately predict molecular properties and structure.
% Existing algorithms rely on single graph representations, typically the minimal free-energy structure,
% neglecting conformational changes even at thermodynamic equilibrium.
% Our approach uses a thermodynamically parameterized exponential family random graph model, i.e., Boltzmann distribution as an embedding.
% We analyze our embedding on the dynamics-driven task of aptamer-ligand binding using experimental data with observed anomalies arising from experimental biases.
% Our results demonstrate that our embedding enables robust community detection and subgraph-level explanations.
% These results enable future experimental design with reduced risk in amplification errors.
% applications in
% \tcb{the design of competent binders
% }
% \tcg{Is this supposed to be completely specific to aptamers or more general something-ligand binding?} % aptamers
% aptamer-ligand binding
% rely on dynamic structural properties and require novel graph embedding methods.
% that impact interactions such as aptamer-ligand binding
% To account for these effects, we embed a Boltzmann-weighted structural ensemble and
% show the equivalence to an exponential family random graph model.
\end{abstract}


\section{Introduction}\label{sec:intro}

Biomolecule graph embeddings built from pairwise intermolecular interaction graphs underpin recent advances in biochemical machine learning, including in the prediction of 
% protein folding~\cite{jumper2021highly}, {RNA family} discovery~\cite{nawrockiInfernal11100fold2013}, and DNA origami~\cite{rothemundFoldingDNACreate2006}.
RNA localization~\cite{akbariroknabadiLGLocNewLanguage2025},
family classification~\cite{rossi2019ncrna},
and binding affinity~\cite{maticzkaGraphProtModelingBinding2014}.
Most graph representations involve single lowest energy biomolecule representations as input. However, single structures fail to capture thermodynamic conformal ensembles, which remain underexplored, especially for weakly folded and dynamic structures such as single-stranded DNA aptamers in solution.
Aptamers--biomolecules with high affinity and specificity for their targets--have a growing impact on biosensing \cite{8207631, nakatsuka2018aptamer}, therapeutics \cite{zhou2017aptamers}, and molecular engineering \cite{liu2006smart}.
% {The design of new aptamer candidates for use in particular applications remains an open challenge.

%Identifying aptamers that perform as needed for real applications remains nontrivial given the inability to fully investigate the hundreds to thousands of candidates identified during selections.

Experimental aptamer selection methods, such as Systematic Evolution of Ligands by Exponential Enrichment (SELEX)~\cite{ellington1992selection}, generate many candidates through multiple rounds of in vitro evolution.
{In this process, selection is tied to sequence abundance, so the final aptamer counts serve as surrogates for binding affinity.} 
Typically, only a small number of candidates can be tested experimentally due to laboratory costs and manual labor. Traditionally one might test only those candidates with highest counts, however 
this indicator is influenced by experimental biases, most notably PCR bias~\cite{takahashi2016high},
% , polymerase errors~\cite{hoinka2012identification}, and self-dimerization~\cite{molecules24193598}
which can leave large numbers of low-count candidates unexplored and contribute to selection failures.
This motivated us to develop enhanced chemically-informed graph embeddings for aptamers to direct the characterization of SELEX-derived aptamer candidates and to improve selection outcomes.

\begin{figure*}[!ht]
    \centering
    \includegraphics[width=1\linewidth, clip, trim = 0cm 0cm 5cm 0cm]{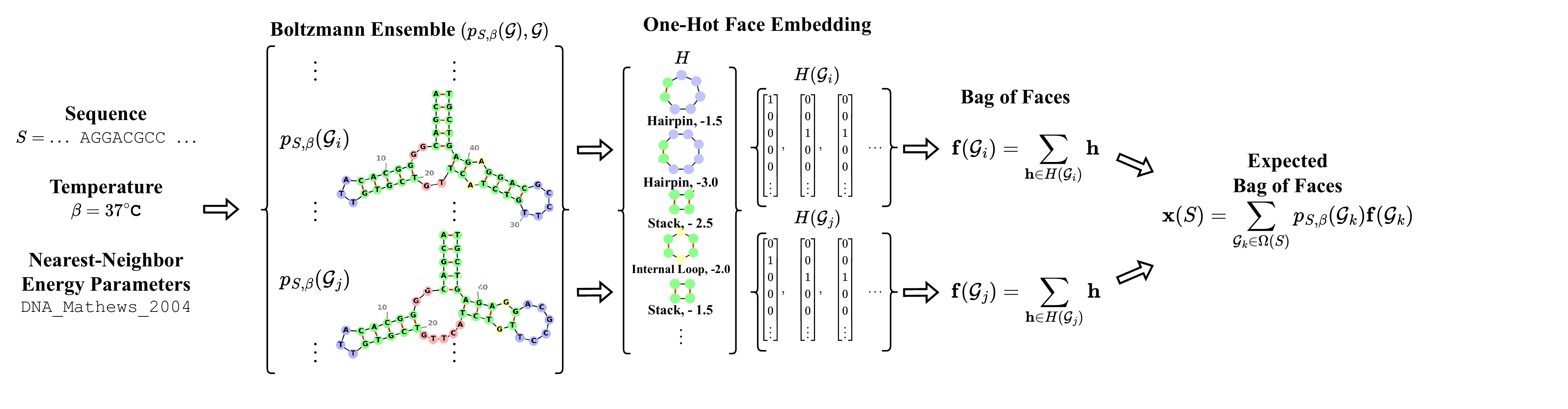}
    \caption{Chemically-informed aptamer embedding via secondary structure ensembles.
    From a sequence of nucleotides, we identify the distribution of secondary structure graphs. Each graph is mapped to a bag of face vector (defined below). 
    Our final features are expected bag of faces vectors defined as sums weighted by the probability distribution. 
    }

    % From a sequence of nucleotides, we identify the distribution of secondary structure graphs. Each graph is mapped to a bag of face vector. We obtain an expected bag of face vector by using the probability distribution as a weighted average. Thus our approach may be viewed as an embedding based on the statistical physics.
    \label{fig:exp_bof_pipeline}
\end{figure*}

%The unique properties of aptamer intermolecular pairwise interactions can be represented by {distinctive graphs, enabling efficient analysis algorithms.}
The primary structure of a DNA aptamer is an oligonucleotide sequence \({S=s_1s_2\dots s_n}\)
where each element $s_i \in \Sigma= \{A,C,G,T\}$ represents one nucleotide base: adenine, cytosine, guanine, or thymine \cite{gmfold2025}.
This primary sequence may be represented by an ordered path $\gG_\text{path} := ( \gV, \gE_\text{path}, \lambda_\gV )$ with node labels $\lambda_{\gV,i}=s_i$.
Secondary structure prediction (folding) is the process determining the edges $\gE_\text{pair}$ that describe pairwise interactions as a partial matching on the path.
For DNA aptamers, the pseudoknot-free behavior of pairwise interactions is equivalent to a non-crossing condition, i.e., there are no pairs $(i,j), (k, l) \in \gE_\text{pair}$ with $i<k<j<l$.
An aptamer's secondary structure is a graph $\gG_\text{fold} :=( \gV, \gE, \lambda_\gV, \lambda_\gE )$
with edges $\gE = \gE_\text{path} \bigcup \gE_\text{pair}$ annotated by $\lambda_\gE \in \{\text{path},\text{pair}\}$. 
% The edges in $E_\text{pair}$ divide the graph into faces, one of which is unbounded. % I don't think this helps in any meaningful way
Importantly, $\gG_\text{fold}$ is outerplanar, meaning that it represents a crossing-free planar embedding
where all vertices lie on the outer face %~\cite{heath1985bookembedding}
% with no vertex strictly inside any cycle~\cite{heath1985bookembedding}
which enables the use of algorithms that are intractable on general graphs~\cite{syslo1983outerplanar}.

% \tcb{Biochemical graph embeddings of are often reliant on their secondary structures.}
The graph representing the minimal free energy (MFE) secondary structure can be found using the Zucker-Stiegler algorithm~\cite{zuker1981optimal} that utilizes dynamic programming (DP).
% to determine the minimum free energy (MFE) secondary structure of an aptamer.
This algorithm identifies the secondary structure that best minimizes the sum of the face energies of the graph (see Sec. II B).
Embeddings of the MFE structure and the face energies have been recently used to successfully cluster similar aptamers~\cite{gmfold2025}.

Single $\gG_\text{fold}$ representations of DNA aptamers are generally insufficient for capturing an aptamer's conformational flexibility\cite{PalLevy2019}, and biologically relevant behavior is often governed by the transition between low-energy conformations\cite{autiero2018intrinsic}.
The Boltzmann distribution describes the secondary structure distributions of an aptamer in equilibrium solution weighted by its thermodynamic properties.
Introducing
partition-functions to DP~\cite{mccaskill1990equilibrium} yields a Boltzmann-weighted ensemble over suboptimal graphs.
This assigns base-pair probabilities $p_{ij}$ and unpaired probabilities $p_{i}$ with $p_i + \sum_k p_{ik} = 1$.
A structural ensemble can then be generated via deterministic backtracking.
% We use these ensembles to construct a physically informed aptamer embedding.

% \tcb{The importance of physics-informed landscape.}
% \tcr{Talk about the ways that people featurize for physics.}
% \tcb{The Boltzmann distribution can be viewed as an Exponential Random Graph Model (ERGM) wherein an exponential function of graph properties determines probabilities corresponding to structures in the ensemble.}
% The Boltzmann distribution can be viewed as an Exponential Random Graph Model (ERGM) wherein 
Exponential-family random graph models (ERGMs)~\cite{robinsIntroductionExponentialRandom2007} provide a modeling method for graph ensembles.
% \marginpar{\tcr{upd. fig1}}
ERGMs use sufficient statistics to compress each graph into motif counts and weights to score those counts, determining which graphs the ensemble prefers.
In our setting, the Boltzmann distribution of pseudoknot-free scondary-structure graphs is an ERGM~\cite{wainwright2008graphical} 
% potential citations:
% \cite{grazioli2019network} explicitly cites a connection between ergms and the boltzmann distribution's states
% https://journals.aps.org/pre/abstract/10.1103/PhysRevE.70.066117 (i don't have access through my institution but the abstract makes it seem promising)
% 
with weights fixed by the thermodynamic parameters and selectable features.

\subsection{Our Contribution}

% \tcr{[include figure ref and adjust slightly to figure description]}
Our work explores embedding aptamer secondary-structure ensembles for anomaly detection. In particular,

\begin{enumerate}
    \item %New embedding %ebof/eneighborhood (can mention both)
    We specify two task-aligned motifs--faces and rooted neighborhoods--and use their expected appearance over the ensemble as embedded feature vectors.
    \item % Aptamer Application
    We apply these embeddings to 
    % high-throughput
    SELEX data, and show that they cluster structurally similar aptamers.
    \item %Aptamer anomaly detection
    We analyze how the embedding space relates to anomalies, enabling community detection against negatives and flagging clusters of anomalous sequences.
\end{enumerate}

Figure \ref{fig:exp_bof_pipeline} illustrates our embedding pipeline to obtain an aptamer's face-type fingerprint.
Starting from a sequence of nucleotides, we identify the
Boltzmann distribution of pairwise interaction probabilities and sample an ensemble of secondary structure graphs.
For each graph in the ensemble, we create a {face-type fingerprint} by summing over the one-hot encodings of its faces, yielding a frequency vector called a bag-of-faces~\cite{gmfold2025} that records the count of each face. 
% \tcr{[Should we use the bag of face terminology without defining it w.r.t. bag of words?]}
We obtain the expected fingerprint of the sequence by computing the mean fingerprint using the probabilities of the Boltzmann distribution. 
{We make our results publicly available at~\cite{ergm-ensemble-embedding}. 
}
% \url{https://github.com/Baker-Data-Science/ergm-ensemble-embedding} .

\subsection{Related Work}

Previous work on SELEX typically embeds candidates with sequence features or single MFE graphs and then clusters structural families; GMFold~\cite{gmfold2025} exemplifies this pipeline, coupling MFE-derived face fingerprints for clustering and similarity search.
ERGMs have been used to study feature-based signatures in graph ensembles~\cite{berlingerio2013network}.
Our work uses ensemble-weighted graph fingerprints for community detection and anomaly analysis, capturing similarity while averaging over ensemble variability.

\section{Background}\label{background}

% This section defines ERGMs and our sufficient statistics.

\subsection{Exponential Family Random Graphs (ERGMs)} 
\label{randomgraphs}
For a fixed sequence $S$, let $\Omega(S)=\{\gG_1,\dots, \gG_m\}$ be the set of all possible pseudoknot-free secondary structures that obey standard base pairing rules.
{ERGMs} provide a generalized framework for defining probability distributions on $\Omega(S)$, 
using a vector of sufficient statistics $\bm{\phi}(\gG)$--for example, face-type and rooted neighborhood counts.
The probability of observing a particular graph $\gG_i\in\Omega(S)$ is given by 
\begin{equation}
p_{S,\boldsymbol{\theta}}(\gG_i) 
= \frac{\exp\!\left( \boldsymbol{\theta}^\top \bm{\phi}(\gG_i) \right)}
{Z_S(\boldsymbol{\theta})}
\end{equation}
where $\boldsymbol{\theta} = (\theta_1, \dots, \theta_k)$ is a vector of real parameters controlling the weight of each statistic and $Z_S$ is the normalizing constant \cite{newman2018networks}
\begin{align*}
Z_S(\boldsymbol{\theta}) = \sum_{\gG\in\Omega(S)}\exp\left( \boldsymbol{\theta}^\top \bm{\phi}(\gG) \right)
\end{align*}
so that $\displaystyle\sum_{\gG\in\Omega(S)}p_{S,\boldsymbol{\theta}}(\gG)=1$.

% \tcb{
% For an ERGM on \(\Omega(S)\) with sufficient statistics \(\bm{\phi}(\gG)\) and parameters \(\boldsymbol{\theta}\), the log-partition function \(\psi_S(\boldsymbol{\theta})=\log Z_S(\boldsymbol{\theta})\) governs all ensemble moments: \(\mathbb{E}_{\boldsymbol{\theta}}[\phi_k(\gG)]=\partial \psi_S(\boldsymbol{\theta})/\partial \theta_k\). In particular, if \(\bm{\phi}\) includes edge-indicator terms \(A_{ij}(\gG)=\mathbf{1}_{\{(i,j)\in E(\gG)\}}\), then the base-pair marginal is \(p_{ij}=\mathbb{E}_{\boldsymbol{\theta}}[A_{ij}]=\partial \psi_S(\boldsymbol{\theta})/\partial \theta_{ij}\). 
The exact evaluation of \(Z_S(\boldsymbol{\theta})\) is generally intractable because the sum ranges over exponentially many structures, and therefore ERGM inference typically relies on approximations (e.g., MCMC-based likelihood or pseudo-likelihood). We later exploit our aptamer-specific structure to enable the rapid computation of $Z_S(\theta)$.
In particular, we utilize DP algorithms with partition functions to recover a probability distribution over admissible structures.
% }

% For a pseudoknot-free secondary-structure ensemble with DP-decomposable energies, the partition function enables us to compute the marginals on the edges via DP.
% Computing the normalizing constant $Z_S(\boldsymbol\theta)$ is computationally prohibitive because it sums over exponentially many folds $\gG\in\Omega(S)$.
% % \tcr{
% However, using deterministic backtracking, we can compute a set of suboptimal structures using an energy threshold $\varepsilon$ and their probabilities.
% \textcolor{red}{[is it not the case that we treat the partition function as $Z_S$ to get the probabilities?]}
% We use a threshold of $\varepsilon=5$ sampling on average $12$ structures.
% \marginpar{\tcr{double-check}}
% }
% \tcr{We therefore restrict our sufficient statistics to those that decompose along the DP, which makes the ERGM view practical for our graphs.}

\subsection{Subgraph Motifs}
\label{planargraph}

We consider two motifs: faces and rooted neighborhoods.

% Let $G=(\gV, \gE, \lambda_{\gV}, \lambda_{\gE})$ be an outerplanar graph.

% \subsection{Outerplanar Graphs}
\paragraph{Faces}
\label{faces}

Let $\gG$ be a graph consistent with Section~\ref{sec:intro}, whose planar embedding partitions the plane into connected open regions, called \emph{faces}. 
The bounded faces are interior faces $\gF_\text{int}$, and the unbounded face is the exterior face $f_\text{ext}$.
% The \emph{exterior face} $f_{\mathrm{ext}} \in \mathcal{F}$ is the unique unbounded face, given by
% \[
% f_{\mathrm{ext}} = \mathbb{R}^2 \setminus \bigcup_{f \in \mathcal{F}_{\mathrm{int}}} f,
% \] \cite{heath1985bookembedding}.
The five categories of aptamer interior faces $f\in\gF_\text{int}$ are defined as
\begin{itemize}
    \item A \textbf{stack} if $(i,j), (i+1, j-1) \in \gE_\text{pair}$.
    \item A \textbf{hairpin} if $(i,j)\in\gE_\text{pair}$ and $\nexists (k,l)\in\gE_\text{pair}$ with $i<k<l<j$.
    \item An \tbf{internal loop} if $\exists \ (i,j), \ (k,l) \in \gE_\text{pair}$ such that $k > i + 1$ and $l < j -1$.
    \item A \tbf{bulge} if $\exists \ (k,l) \in \gE_\text{pair}$ such that $k=i+1, \ l < j-1$, or $k > i+1, \ l = j-1$.
    \item A \tbf{multibranch} if $\exists \ (k,l), \ (k', l') \in \gE_\text{pair}$ such that $k' > k+1$ and $l' < l-1$.
\end{itemize}

Each face has an empirically measured free energy $E(f)$ where the total energy is an additive sum over the faces $E(\gG)=\sum_f E(f)$.
Recent graph-based work makes this explicit: the faces of the secondary-structure graph are taken as fundamental objects with associated energies, enabling fast subgraph/face matching across sequences~\cite{gmfold2025}.
% This face energy 
% \tcr{add more about the connection to energy here}

\begin{figure}[!ht]
    \centering
    \includegraphics[width=1\linewidth]{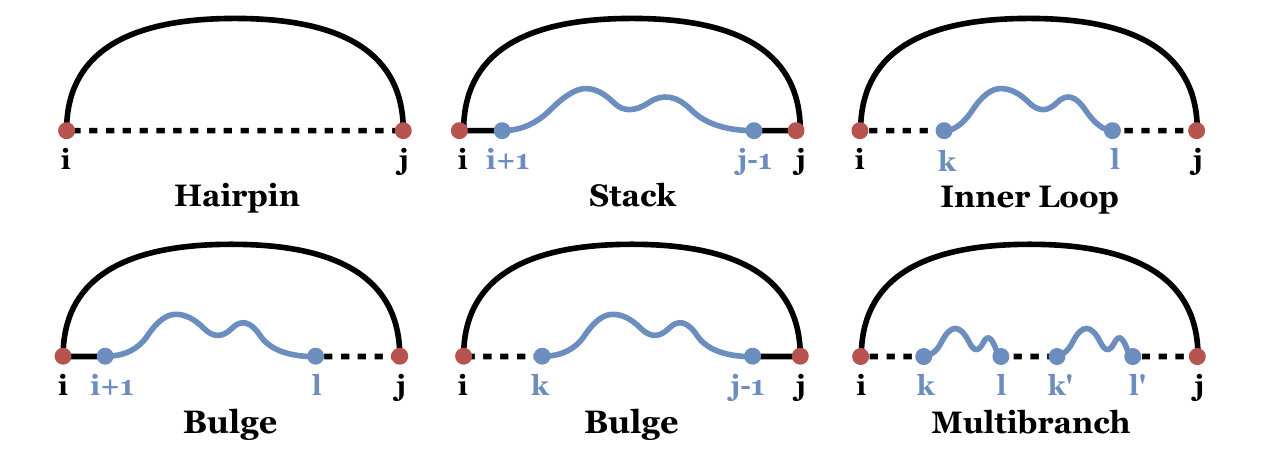}
    
\caption{Illustration of the five standard aptamer face types: hairpin, stack, internal loop, bulge and multibranch.
Each face has a defining edge $(i,j)\in\gE_\text{pair}$ shown by a black arc. The path between $(i,j)$ is represented by a dashed line and any nested regions bounded by an edge in $\gE_\text{pair}$ are illustrated by blue arcs.
In addition to type, each face has an associated energy depending on its nucleotide makeup. We consider (type, energy) pairs as subgraph motifs, as presented in~\cite{gmfold2025}.
}
\label{fig:facetypes}
\end{figure}

% \marginpar{\tcr{upd. fig2}}

Figure \ref{fig:facetypes} illustrates a categorization of $\gF_\text{int}$ into stack, hairpin, internal loop, bulge, and multibranch.
Every ${f\in\gF_\text{int}}$ has a defining edge $(i,j)\in\gE_\text{pair}$ represented by a black arc. The path between $i$ and $j$ is shown by a dashed line, and any nested regions bounded by an
edge in $\gE_\text{pair}$ are illustrated by a blue curve. 
Each $f\in\gF_\text{int}$ is also assigned an energy $s(f)\in\sR$ by the DP folding algorithm for the chosen energy model. In our embedding, we ignore vertex counts and nucleotide labels and count faces according to their (type, energy) pair, consistent with~\cite{gmfold2025}. 

% \subsection{Rooted Neighborhood}
\paragraph{Rooted Neighborhoods}
\label{rootedneighborhood}
We may incorporate subgraphs as features by using rooted neighborhoods as motifs.
First, we consider the radial distance on the graph $d_\gG(u,v)$, the minimum number of edges required to travel between nodes $u,v\in\gV$. Fixing a \emph{central node} $c\in \gV$ and a \emph{radius} $r\in \mathbb{N}_0$, the closed radius-$r$ rooted neighborhood of $c$ is
\[
N^\gG_r(c)\;:=\;\{\,v\in \gV:\ d_\gG(v,c)\le r\,\}.
\]
Importantly, $N^\gG_r(c)$ induces a subgraph on $\gG$ denoted $\gG[N_r(c)]$.
Given the outerplanar structure of the initial graph, for sufficiently small $r$ the graph isomorphism problem for all $N^\gG_r(c)$ is computationally feasible~\cite{Costa2010FastKernel}.
In our embedding, we count the isomorphic radius-$r$ rooted neighborhoods.

% \subsection{Dual Graphs}
% \label{dualgraph}
% The set of faces $\mathcal{F}$ in a planar graph can be recontextualized to construct a dual graph representation in which every vertex corresponds to a face in the original graph. Given a planar graph $G$ with a set of faces $\mathcal{F}$, the dual graph $G^*=(V^*, E^*, \lambda_{V^*}, \lambda_{E^*})$ is defined by associating a vertex $v^*\in V^*$ with every face $f \in \mathcal{F}$ and an edge $e^*$ with every edge $e \in E$ that is adjacent between two faces in the planar embedding of $G$ \cite{nishizeki1988planar}.  This interpretation enables more compact graph representations of structural features in the original planar embedding, facilitating simple/...straightforward graph embeddings that preserve local and global structural properties..... 

\section{Statistically Informed Fingerprints}
\label{sec:approach}

% We have established the utility of several notions relating to graphs, such as outerplanar graphs, neighborhoods, and ERGMs. However, we have also been laying the framework for a larger problem that requires a solution that combines these techniques. For highly dynamic networks characterized by volatile outerplanar exponential graphs, the underlying probabilistic distribution can inform the structural feature fingerprint to produce enhanced embeddings. This new embedding not only allows us to mathematically capture the dynamic nature of graphs, but allows flexibility in construction. 
% \subsection{Fingerprint Embedding}
% Aim to formulate a less programming-based method of creating a) the library of faces collected from all aptamers and b) the individual bag of faces vector:
% h_i \in \mathcal H <— space of all possible “features” so you use this for your N^{|\mathcal H|}
% H = {(h_1, c_1), (h_2, c_2), …} set of feature counts for one graph in one ensemble
We develop an embedding designed to reflect aptamer structural flexibility for more informed feature analysis.
% \tcr{Show that the Boltzmann distribtuion is an ERGM.} is this strictly necessary? -Alex
\subsection{Boltzmann Distribution as an Aptamer ERGM}
% \tcb{
Let \(\Omega(S)\) be the set of pseudoknot-free secondary-structure graphs for sequence \(S\). For \(\gG\in\Omega(S)\) with free energy \(E(\gG)\) (kcal/mol), the Boltzmann ensemble
with Boltzmann constant $k_B$
at temperature \(T\) is
% \[
% p_{s,\beta}(\gG)=\frac{\exp(-\beta E(\gG))}{Z_s(\beta)},\quad
% Z_s(\beta)=\sum_{G'\in\Omega(s)} \exp(-\beta E(G')), \quad
% \beta=\frac{1}{RT}.
% \]
$$p_{S,\beta}(\gG) = \frac{\exp\left(-\beta E(\gG)\right)}{Z_S(\beta)},\quad 
Z = \sum_{i=1}^{m} {e^{-\beta E(\gG)}},\quad 
\beta = \frac{1}{k_BT}.
$$
% with 
% $$
% Z = \sum_{i=1}^{m} {e^{-\beta E(\gG_i)}},\quad 
% \beta = \frac{1}{k_BT}
% $$
If \(E(\gG)=\sum_k w_k\,t_k(\gG)\) for motif counts \(t_k(\gG)\), then \(p_{s,\beta}\) is an ERGM with sufficient statistics \(t(\gG)\) and parameters \(\theta_k=-\beta w_k\), i.e., \(p_{s,\beta}(\gG)\propto \exp(\theta^\top t(\gG))\).\\ 
% The base-pair and unpaired marginals are
% \[
% p_{ij}=\sum_{\substack{\gG\in\Omega(S)\\(i,j)\in E(\gG)}} p_{s,\beta}(\gG),
% \qquad
% p_i=\Pr[i\ \text{unpaired}].
% \]

In practice, \(Z_s(\beta)\) and \(\{p_{S,\beta}\}\) are computed exactly via ViennaRNA~\cite{wuchty1999complete}.
Energies are expressed in kcal/mol. The Boltzmann constant $k_B=1.98\times 10^{-3}\textrm{kcal}\ \textrm{mol}^{-1}\ \textrm{K}^{-1}$. We use a temperature of $37^\circ$ C (310.15 $K$) which is consistent with the temperature of the data collected in the SELEX experiment. 
To compute DNA energies, we use  the set of DNA nearest-neighbor energy parameters $\textsc{DNA\_Mathews\_2004}$~\cite{matthewsOrderedDNARelease2004}. %The algorithm applies dynamic programming to evaluate the minimum free energy achievable in each substructure under the nearest-neighbor model, before using deterministic backtracking to enumerate all suboptimal structures and free energies within a specified threshold above the minimum free energy.

\subsection{Statistically Informed Fingerprints}
% Second, we show how this can be paired with graph \tcr{fingerprints} to build an expected graph fingerprint.
We represent each sequence $S$ by the ensemble expectation of its graph features.
Let $H=\{h_1,\ldots,h_d\}$ be a global feature dictionary (either faces or rooted neighborhoods).
For any fold $\gG\in\Omega(S)$, we define the feature-count vector $\vf(\gG) \in \mathbb{N}^d$ by its entries 
$\vf_k := \#\{\, h_k \;\text{occurs in } \gG \,\}$.
Given the Boltzmann ensemble $p_{S,\beta}(\gG)$, the ensemble-weighted fingerprint 
% for $S$
is
\[
\vx(S) = \mathbb{E}_{\gG\sim p_{S,\beta}}[\vf(\gG)] = \sum_{\gG\in \Omega(S)} p_{S,\beta}(\gG) \vf(\gG).
\]

We remark that the feature vector $\vf(\gG)$ is a
bag-of-faces~\cite{gmfold2025}
% \textcolor{red}{[this is the first time 'bag of face' term has been used. if we need to briefly introduce the concept or that it's inspired by bag of words we could either do that here or insert it in related work with GMFold]}
or bag-of-neighborhoods.
We refer to our ensemble-weighted feature vector $\vx(S)$ as an expected bag-of-faces or expected bag-of-neighborhoods.
% We denote our embedding an expected bag-of-face or expected bag-of-neighborhoods.
Moreover, the dimension of the dictionary can be exceptionally large for all subgraphs.
As a result, we maintain a relatively low $r=4$.
Additionally, as these are expected counts, the resulting feature vectors are non-negative, making the embedding useful for techniques such as non-negative matrix factorization (NMF)~\cite{NIPS2000_f9d11525}.
\label{statsfingerprints}
\label{expectedfaces}
%\Faces of an outerplanar graph fall into four categories: the first of these is a hairpin loop, which is a subgraph with one interior edge. An inner loop is a subgraph with two nonadjacent interior edges, while a bulge is a subgraph with two adjacent interior edges. A multi-branch is a subgraph that contains more than two interior edges, and finally a stack is characterized by two interior edges surrounded by one exterior edge on both sides. \cite{zuker1981optimal} \cite{wiegreffe2019rnapuzzler}
%Let G = $(V,E,\lambda_V,\lambda_E)$ be an outerplanar graph. Let $(i,j) \in E_\text{int}$ such that $i \in V$ and $j \in V$ are the starting and ending node delimiting a face.
%Let U = $\{i+1, \dots, j-1\}$ be the set of nodes between $i$ and $j$. Let $U_\text{int} = \{(k, l) \in E_\text{int} \; | \; i < k < l < j\}$ denote the set of interior edges belonging to a set of interior edges between $i$ and $j$.

%\subsubsection{Expected Rooted Neighborhoods}

%The second feature we use is rooted neighborhood, as defined in \ref{rootedneighborhood}. For each node $x$ in a graph, we consider the neighborhood centered at $x$ with radius $r$. Each graph is embedded into a vector that records the number of times a particular neighborhood appears. Similar to the Bag of Faces model, the final vector contains an analogous bag of subgraph neighborhoods based on the ERGM's probability distribution.

\section{Applications to Aptamers}

We process and partially label anomalous SELEX data and apply community detection on our embeddings. Seven experimentally tested high-binding aptamers are used as validation.
% We analyze our embedding in both supervised and unsupervised machine learning environments. 

\subsection{Processing SELEX Data} % Results?
% \tcr{SELEX is an iterative enrichment process, where aptamers that bind to target are preferentially amplified and carried into subsequent rounds --> Do we need to restate this? }
We utilize unprocessed SELEX data from~\cite{gmfold2025} consisting of next-generation sequencing (NGS) of aptamer candidates against the target norepinephrine. There are two libraries with two rounds N48: 9 \& 13, and N58: 12 \& 16.
For each library-round $(l,r)$ we observe sequence-count pairs
% Our experimental data consists of two aptamer libraries, N48 and N58, with two rounds of SELEX in each, totaling 3711 unique aptamers after removing mistranscriptions.
% $$R^{(l,r)} = \left\lbrace(S^{(l,r)}_{i}, C^{(l,r)}_i) \mid i \in \{1, 2, \dots,n \}\right\rbrace$$
$(S^{(l,r)}_i,C^{(l,r)}_i)$
where $S^{(l,r)}_i$ is aptamer $i$'s primary sequence and $C^{(l,r)}_i$ its SELEX read count.

% Our quality filtering differs from~\cite{gmfold2025}, which considered the maximum count over all of the multisets for a given sequence: 
Viewing SELEX as a dynamic process, we remove any aptamers that emerge in a later round without being in a prior round, which we observe as mutations with low counts.
This leaves 3711 unique aptamers across all of the libraries and rounds.
Among these, six aptamers exhibit an experimentally validated high binding affinity~\cite{mitchell2025advancing}.
% \textcolor{red}{(nitpick) technically one of those seven wouldn't be 'among these' since it was a mistranscription.}
% \tcr{also indicate known good binder}

% The aim of SELEX is to use the aptamer counts in the final round as a surrogate for binding affinity.
% The SELEX count distribution exhibits a high degree of kurtosis, where there is a large peak centered at low counts and then very few exceptional outliers.
Each unique aptamer is assigned a count based on the last round in which it appears.
Counts are normalized in each library.
% in each library is normalized before normalizing across libraries.
We map each count $c_i$ to a unit-interval score by assigining it to decile intervals $I_j=[\frac{j}{10},\frac{j+1}{10}]$. Within each bin $j$, items are positioned at evenly spaced midpoints  yielding uniform spacing inside each decile.
% The final count 
% the final aptamer count using \tcr{count-per-million}
% \[
% c^{(l)}_{i} = log_2 \left( \frac{C^{(l,y)}_i \times 10^6 + 1}{\sum_{j}C^{(l,y)}_j}\right). 
% \]
% \textcolor{red}{Then, we normalize across libraries by averaging over the total number of aptamers. [I am not sure what this is referring to?]} 
If an aptamer appears in both libraries, we consider the maximum CPM over the libraries as its final CPM, $c_i = \max\{c_i^{(N48)},c_i^{(N58)}\}$.

The advantage of utilizing multiple round data is that it provides notions of trend in the SELEX process.
We call our trend metric selective pressure, which measures an aptamer's change in count between rounds, defined by
% We define these metrics as
% \tcr{no enrichment? I think we need enrichment for the nonlinear part no? we do. Ok we should add that back}.  
\[
{\rho}^{(l)}_{i} = \frac{C^{l,y}_i - C^{l,x}_i}{C^{l,x}_{i}}
\]
%SHOULD THIS BE C^{l,x}_i in the denominator?
where we have round $x<y$. If an aptamer appears in both libraries, its total pressure is the sum of its pressures from each library, $\rho_i = \rho^{(N48)}_i + \rho^{(N58)}_i$.

% In the SELEX process, it is possible that competitive aptamers go unobserved due to the significant kurtosis of the distribution.
Using counts and trend, we partially label anomalous aptamers as those over valued by count, i.e., high-count low-pressure (HC-LP) and those under valued by count, i.e., low-count high-pressure (LC-HP).
We use thresholding to determine over-valued anomalies $S^{\text{HC-LP}}$ and under-valued anomalies $S^{\text{LC-HP}}$
\begin{align*}
S^{\text{LC-HP}}&=\{ S_i \in \{1, 2, \dots,n \} \mid c_i \leq c^*, \rho_i \geq \rho^*\}\\
S^{\text{HC-LP}}&=\{ S_i \in \{1, 2, \dots,n \} \mid c_i \geq c^*, \rho_i \leq \rho^*\}
\end{align*}
where $c_i$ and $\rho_i$ are counts and pressures, respectively.
We use the 90th percentile for $c^*$ and 10th percentile for $\rho^*$. 
This rank-based threshold is scale-invariant across rounds and aligns with common practice for stabilizing heavy-tailed SELEX data with quality filters~\cite{kohlbergerSELEXCriticalFactors2022}.

% \marginpar{\tcr{upd. fig3}}
\begin{figure}[!ht]
    \centering
    \includegraphics[width=1.0\linewidth]{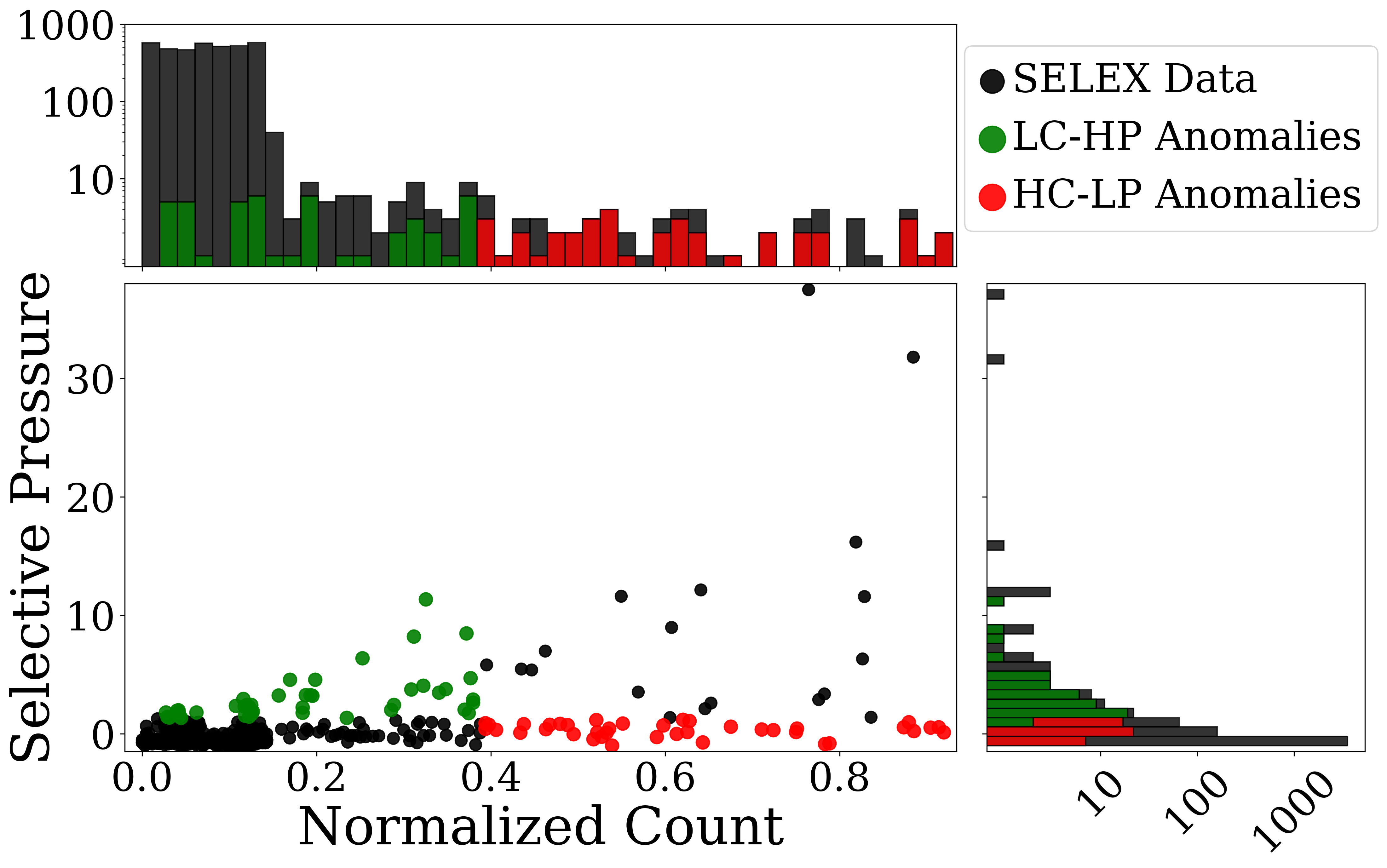}
    \caption{Selective pressure versus count per million (center) with the histograms for normalized count (top) and selective pressure (right).
    Low-count high-pressure anomalies are marked in green and high-count low-pressure anomalies are marked in red.
    % The histograms are log scaled and exhibit strong kurtosis with an extreme peak near 0 and a few exceptional outliers.
   }
    \label{fig:enrich_cpm_stats}
\end{figure}

Figure 3 illustrates the data distribution along with the partially labeled anomalous structures.
In particular, it shows selective pressure versus count per million (center) with the
histograms for count per million (top) and selective pressure (right).
Low-count high-pressure outliers are marked in green and high-count
low-pressure outliers are marked in red. The histograms are log scaled
and exhibit strong kurtosis with an extreme peak near 0 and a few
exceptional outliers.
The red regime contains likely artifacts or exhausted candidates—sequences with high abundance yet little or negative pressure, consistent with amplification bias.
The green regime contains emergent aptamers that are still under-represented but show large round-to-round gains and may be early binders.

% \section{Results}
% \begin{enumerate}
%     \item t-SNE plot with clusters
%     \item Random forest plot with clusters

% \end{enumerate}
% \subsection{Physically Informed Graph Feature Analysis}

% Count-based methods of aptamer candidate selection are vulnerable to the experimental imperfections associated with SELEX, reliant on data that misrepresents true aptamer quality. We propose a hypothesis generation strategy that is resistant to this distortion by incorporating adversary detection. Adversaries in this context are low-performing aptamers with artificially elevated counts, representative of the fallibility of count data. Adversary detection can be performed by identifying the graph features that are correlated with low selective pressure and clustering the dataset according to similarity in such features, so that clusters that are likely to contain adversaries can be eliminated from consideration. 

Our embeddings are constructed by concatenating the expected bag-of-faces and expected neighborhoods to a k-mer embedding~\cite{liu2006smart} with $k=4$. The resulting feature matrices are
$\mX_\text{EBOF} \in \mathbb{R}^{3711 \times 3102}$ and $\mX_\text{EN} \in \mathbb{R}^{3711 \times 850}$.
% \textcolor{red}{do we need to justify the choice to concatenate kmer?}
% , with each row $i$ a concatenation of aptamer $i$'s k-mer embedding with its expected features. 

\subsection{Robust Community Detection}

% In this section, we use both Random Forest and XGBoost Classifiers to assign an "importance" to each feature. These ensemble learning methods build decision trees to define complex decision boundaries of a non-convex input space, allowing the model to identify which features contribute the most to classification by computing their "feature importance." 

First, we perform community detection on the embeddings directly, and observe if there are any clusters robust to anomalous aptamers.
We perform topic modeling with NMF with 25 topics on $\mX_{\text{EN}}$--a dimension reduction technique factoring  $\mX \approx \mM\mH$, where $\mM$ attributes topics to data points and $\mH$ attributes features to topics.
We then cluster $\mM$ using spectral clustering~\cite{von2007tutorial} with 35 clusters, chosen by sweeping from 5 to 50 clusters and selecting the highest
% \tcr{higher is better for silhouette}
silhouette score.
For visualization, we perform further dimension reduction via t-SNE~\cite{maaten2008visualizing}.

\begin{figure}[!ht]
    \centering
    \includegraphics[width=0.9\linewidth, clip, trim = 1.cm .75cm 1.cm .7cm]{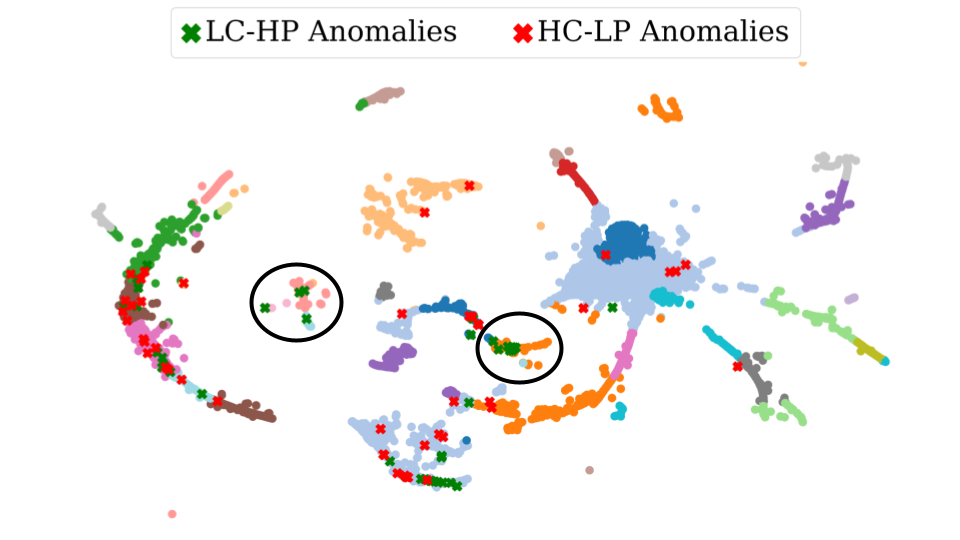} 
\caption{Two-dimensional t-SNE embedding after applying NMF with 25 topics.
Points are colored by cluster using spectral clustering to identify 35 clusters.
Green X's are LC-HP aptamers, red X's are HC-LP aptamers, and
two robust neighborhoods are circled in black.
% The black circle identifies a cluster with no HC-LP anomalies and a statistically significant average enrichment and count per million.
}
  \label{fig:xgb_clustering}
\end{figure}

Figure \ref{fig:xgb_clustering} illustrates the embedding's capacity for isolating robust communities, visualized via a t-SNE embedding on $\mM$.
Clusters are indicated by coloration, with noisy data in translucent gray. Red crosses represent over-valued (HC-LP) anomalies and green crosses represent under-valued (LC-HP) anomalies.
% \marginpar{\tcr{upd. fig4}}
Circled in black are clusters without over-valued
anomalies, which showcases our method's effectiveness in isolating aptamers based on shared binding characteristics. 
These clusters also exhibit the highest average selective pressures and counts per million, indicating potential for high binding affinity.
% \tcr{[No explanation of model used? We originally did clustering here based on feature importance, should we rerun the figure for clustering on the whole set of features?]}

\subsection{Exploring Subgraph Level Explainability}
% \tcr{can we intro tsne for vis once in the beginning instead of each time in fig caption}
Next, we assess the correlation between the embedded features and the selective pressure and its ability to perform community detection using the partial labeling.
To establish correlation, we use a linear model
$\mX_\text{EBOF}\vw = {\rho} $
where $\vw \in \mathbb{R}^{3102}$ is estimated using a Ridge regressor with a least-squares solver.
Table~\ref{tab:features-combined} shows the features with the most negative and positive coefficients. 
% in $\mathcal{\Phi_\text{neg}}$ and $\mathcal{\Phi_\text{pos}}$ with their coefficients given by the Ridge regression model.
%with parameters $\alpha = 100$.

\begin{table}[!ht]
\setlength{\tabcolsep}{.12cm}
\scriptsize
\centering
\begin{tabular}{l r l r}
\toprule
\multicolumn{2}{c}{Most negative} & \multicolumn{2}{c}{Most positive} \\
\cmidrule(lr){1-2}\cmidrule(l){3-4}
Feature & {Coef.} & Feature & {Coef.} \\
\midrule
INTERNAL:9+8:AT/TA      & -0.45 & INTERNAL:6+7:AT/TG & 0.50 \\
ATTC                    & -0.13 & BULGE:17:AC/TG     & 0.42 \\
BULGE:11:CA/GT          & -0.10 & BULGE:11:AA/TT     & 0.25 \\
CATG                    & -0.09 & TTTA               & 0.23 \\
INTERNAL:12+11:AT/TA    & -0.09 & ATTT               & 0.23 \\
\bottomrule
\vspace{0.05in}
\end{tabular}
\caption{Top five negative and positive features by Ridge coefficient
(negative on the left, positive on the right). Faces are typed and
nucleotide-specific; 4-mers are plain strings.}
\label{tab:features-combined}
\end{table}
To use our partial labeling in anomalous community detection, we construct an embedding from all features with negative coefficients:
\begin{align*}
\mW_{-} = \mX_{[:, \mathcal{I}_{-}]} \quad \text{where} \quad \mathcal{I}_{-} = \{ i \mid \vw_i < 0 \}.
\end{align*}
% I think this notation for C might be too code-like because of the slicing
Here community detection is similar, modeling 25 topics of $\mW_{-}$ with NMF.
We use spectral clustering~\cite{von2007tutorial} to cluster $\mM$ into 25 clusters and t-SNE to visualize.

To detect anomalous clusters, we calculate for each cluster a weighted sum of the average cluster coefficients
$$\Delta_C = \frac{1}{5} \sum_{j \in \vw_{\text{neg}}} \bar{\vx}^{(C)}_{j} - \frac{1}{5} \sum_{j \in \vw_{\text{pos}}} \bar{\vx}^{(C)}_{j},\quad \bar \vx^{(C)} = \frac{1}{|C|}\sum_{i\in C}\mX_{i}$$
selecting only $\vw_\text{neg}$ and $\vw_\text{pos}$, the 5 features with the most negative coefficients and the 5 features with the most positive coefficients, respectively. The 10 clusters with the highest $\Delta$ are labeled anomalous $\gC_\Delta = \{i\in C\, \mid \, \Delta_{C} > n\}$.

% \begin{table}[!ht]
% \centering
% \begin{tabular}{|l|r|}
% \hline
% \textbf{Feature} & \textbf{Coefficient} \\
% \hline
% INTERNAL:9+8:AT/TA & -0.4461 \\
% ATTC               & -0.1290 \\
% BULGE:11:CA/GT     & -0.0979 \\
% CATG               & -0.0914 \\
% INTERNAL:12+11:AT/TA & -0.0910 \\
% \hline
% \end{tabular}
% \vspace{1em}
% \caption{Five Most Negative Overlapping Negative and Their Coefficients. The left column shows the name of the feature, face or 4-mer. Every face is defined by its type followed by the nucleotides contained in it. Each 4-mer is a string of 4 nucleotides. The right column shows the coefficient for each feature as estimated by the ridge regression model.}
% \label{tab:negativefeatures}
% \end{table}

% \begin{table}[!ht]
% \centering
% \begin{tabular}{|l|r|}
% \hline
% \textbf{Feature} & \textbf{Coefficient} \\
% \hline
% INTERNAL:6+7:AT/TG & 0.5023 \\
% BULGE:17:AC/TG     & 0.4178 \\
% BULGE:11:AA/TT     & 0.2524 \\
% TTTA               & 0.2339 \\
% ATTT               & 0.2293 \\
% \hline
% \end{tabular}
% \vspace{1em}
% \caption{Five Most Positive Overlapping Features and Their Coefficients. The left column shows the name of the feature, face or 4-mer. Every face is defined by its type followed by the nucleotides contained in it. Each 4-mer is a string of 4 nucleotides. The right column shows the coefficient for each feature as estimated by the ridge regression model.}
% \label{tab:positivefeatures}
% \end{table} 

\begin{figure}[!ht]
        \centering
            \includegraphics[width=1\linewidth]{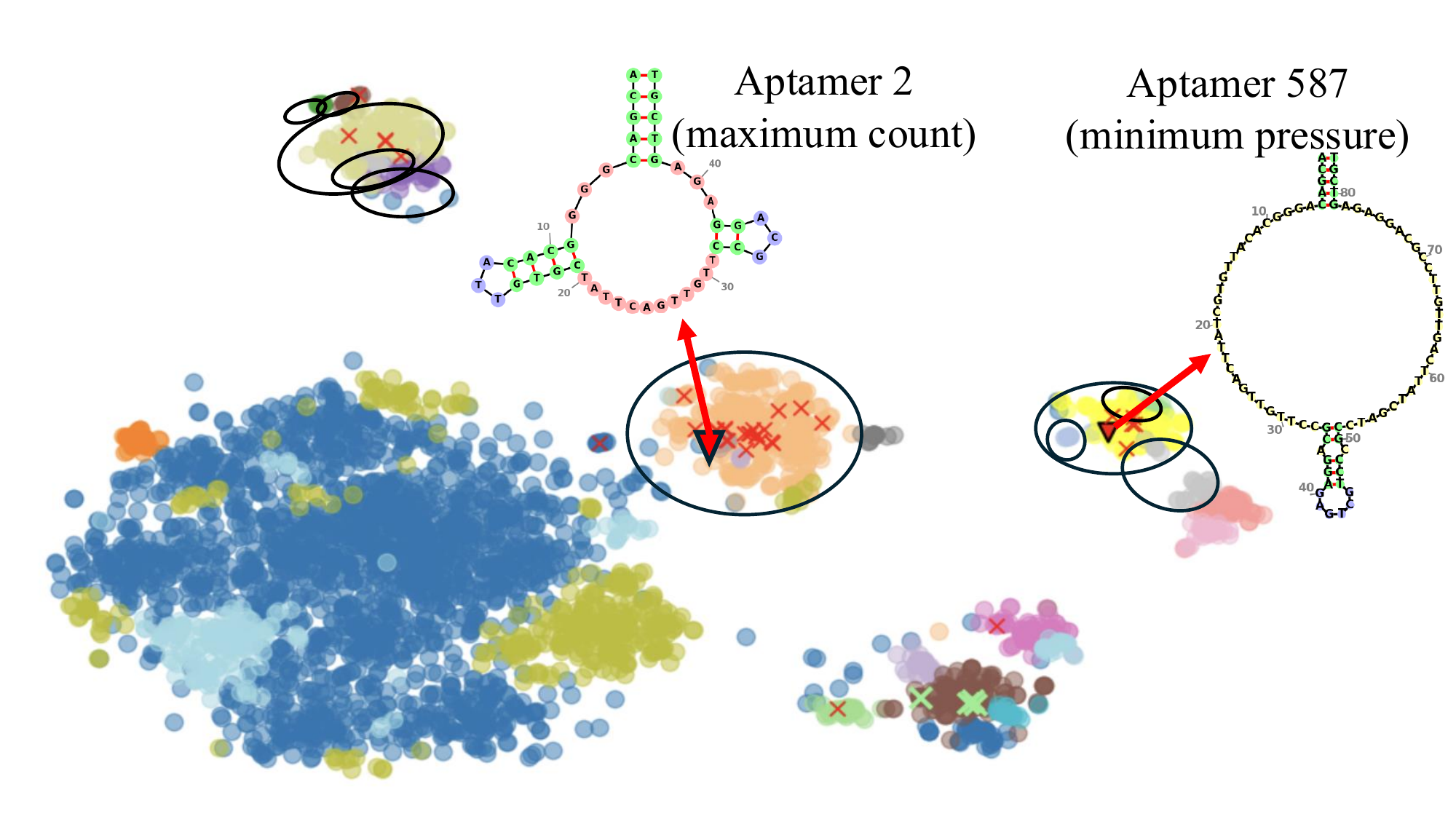}
                \caption{Two-dimensional t-SNE embedding on negatively correlated features. Points are colored by cluster using spectral clustering to identify 25 clusters. Circled in black are identified anomalous clusters. Red X's are over-valued anomalies (HC-LP) and green X's are tested good binders. The structures of two captured over-valued anomalies are shown.}
    \label{fig:tsneadversary}
\end{figure}

Figure \ref{fig:tsneadversary} shows the results for community detection.
% Here, we define adversaries by setting $c^*$ as the 90th percentile cutoff for count and $\rho^*$ as the 10th percentile cutoff for pressure. 
Clusters are illustrated by color and over-valued anomalies are labeled with a red cross.
Anomalous clusters are circled in black. Several of these visibly contain a higher number of over-valued anomalies, showing how the community detection method allows us to discard misleading high counts. For instance, aptamer 2--a high-count low-pressure aptamer--is identified as an over-valued anomaly. We also discard aptamer 587, the aptamer with the lowest selective pressure in the data set. Most tested positive binders (marked with green crosses) are not in a circled cluster, indicating that our method separates over-valued anomalies from high-performing aptamers.

% \usepackage{booktabs,siunitx}
% \sisetup{table-number-alignment = center, table-format=1.4}

We eliminate anomalous clusters to recommend promising aptamers for further testing. In particular, taking
\begin{align*}
\mW_{+} = \mX_{[\gI, \mathcal{I}_{+}]} \quad \quad \mathcal{I}_{+} = \{ i \mid \vw_i \geq 0 \}, \quad \gI = \{i \mid i\not\in\gC_\Delta \},
\end{align*}
we apply the same NMF and t-SNE procedure to $\mW_{+}$.
% To do so, we first group the remaining aptamers by their similarity in positive feature using NMF with 25 topics--a dimension reduction technique factoring $X \approx WH$, where $W \in \mathbb{R}^{3711 \times 25}$ attributes topics to data points and $H \in \mathbb{R}^{25 \times 3102}$ attributes features to topics.

\begin{figure}[!ht]
            \centering
            \includegraphics[width=1\linewidth]{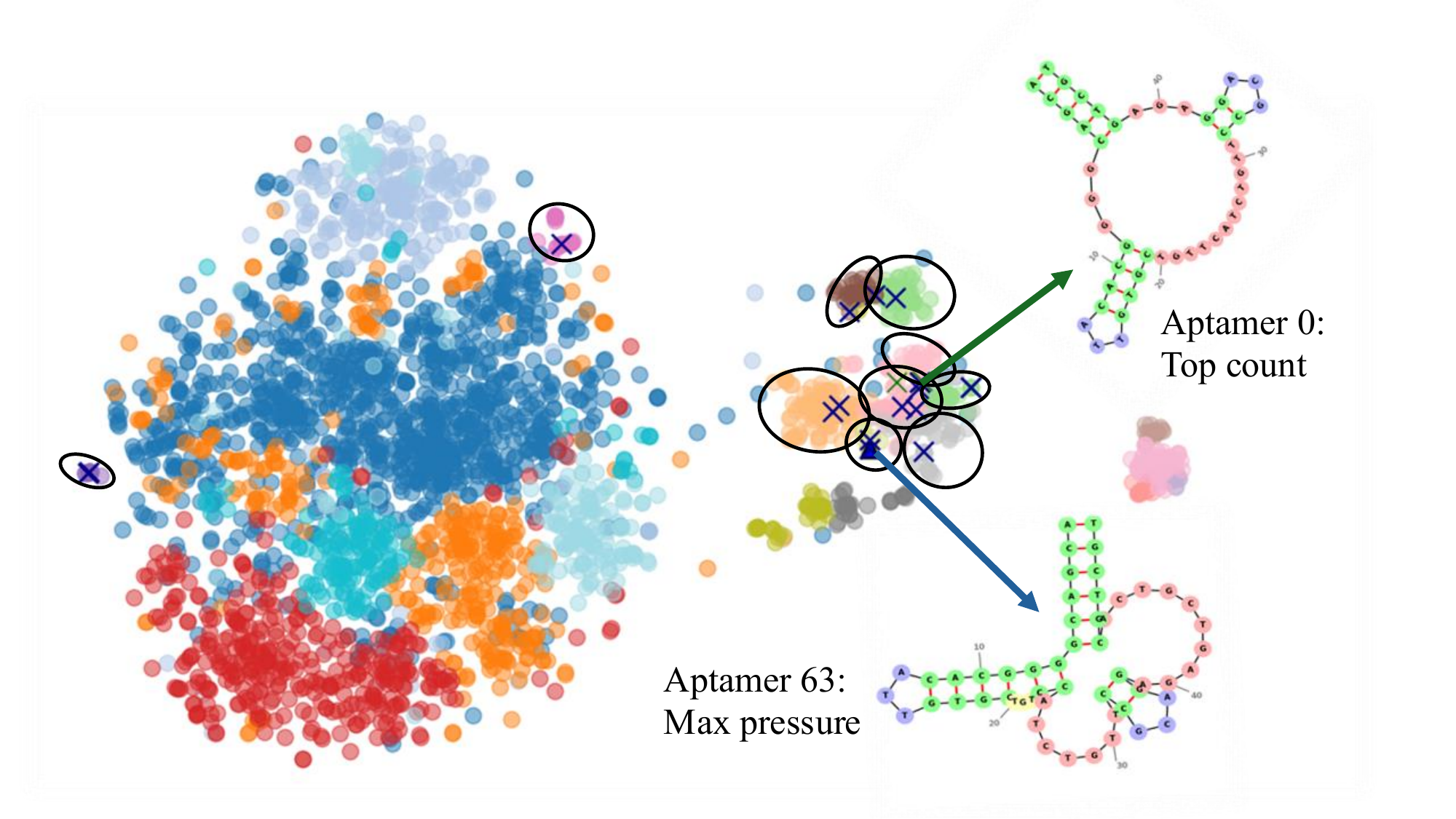}
\caption{Two-dimensional t-SNE embedding on positively correlated features, excluding aptamers from anomalous clusters. Points are colored by cluster using spectral clustering to identify 25 clusters. Recommended clusters are circled in black. Green X's are tested good binders, and blue X's are aptamers we recommend for further testing, i.e., the highest count and highest pressure in each cluster. The structures of two recommended aptamers are shown.
% : aptamer 0, the highest count, and aptamer 63, the highest pressure in the data set.
}
        \label{fig:tsnehypgen}
    \end{figure}

Figure \ref{fig:tsnehypgen}  illustrates the communities of aptamer candidates produced by our hypothesis generation method, visualized with the t-SNE embedding on $\mW_{+}$.
% \tcr{[what's $C'$]}.
Recommended clusters are circled in black, and further downselection can be performed by prioritizing sequences with the highest count and selective pressure of the cluster, corresponding to the blue crosses on the plot. Confirmed good binders, marked by green crosses, are located in two of the clusters we recommended.
% Recommended aptamers include aptamer 0, the highest count of the data set and a confirmed good binder, as well as aptamer 63, the aptamer with the highest pressure of the data set
It is notable that tested good binders and the majority of recommended clusters are placed close together on the right side of the plot, indicating similarity in ${\vw}_\text{pos}$ feature counts. Therefore, in future SELEX runs, building an initial library out of aptamers enriched for features in  ${\vw}_\text{pos}$ and depleted of features in $\vw_{\text{neg}}$ may enhance aptamer discovery. 
% \tcr{Nevertheless, the marked aptamers appear evenly separated into different clusters within that common region, suggesting a degree of feature diversity among the most promising binders $\rightarrow$ do we need to say this? } 

\section{Conclusion}

We propose a chemistry-informed graph embedding that represents aptamers as Boltzmann-weighted ensembles of secondary structures, cast in an exponential-family view with DP-compatible motif statistics.
Applied to two SELEX libraries, the embeddings enable community detection, identification of anomalies, and subgraph-level explanations linking face and neighborhood patterns to selection signals.
Key limitations include the exclusion of pseudoknots and fixed thermodynamic parameters, with future directions in parameter learning, multi-temperature ensembles, and experimental validation.

% We introduced a physics-informed graph embedding that averages over a Boltzmann-weighted ensemble of pseudoknot-free secondary structures, casting the ensemble in an exponential-family (ERGM) view and using DP-compatible motif  (face-type counts and rooted neighborhoods) to form ensemble-weighted fingerprints.
% In two SELEX libraries, these embeddings support community detection, detection of high-count-low-pressure anomalies, and subgraph-level explanations that point to specific loop and neighborhood patterns associated with the selection signal.
% Limitations include the exclusion of pseudoknots, multi-strand/primer effects, and reliance on fixed thermodynamic parameters. Promising next steps are extending the statistic set to include additional temperatures, learning or calibrating ensemble parameters from SELEX, modeling covariances for uncertainty-aware scoring, and validating hypothesized communities with experimental binding measurements.

\bibliographystyle{IEEEtran}
\bibliography{references}
\end{document}